%% file: main.tex
\documentclass[10pt,twocolumn,letterpaper]{article}

\usepackage[pagenumbers]{cvpr}              % To produce the CAMERA-READY version
\input{preamble}
\definecolor{cvprblue}{rgb}{0.21,0.49,0.74}
\usepackage[pagebackref,breaklinks,colorlinks,allcolors=cvprblue]{hyperref}

\hypersetup{hidelinks}
\usepackage{url}

\usepackage{booktabs}       % professional-quality tables
\usepackage{amsfonts}       % blackboard math symbols
\usepackage{nicefrac}       % compact symbols for 1/2, etc.
\usepackage{microtype}      % microtypography
\usepackage{xcolor}         % colors
\usepackage{amsmath}
\usepackage{graphicx}
\usepackage{amsthm}
\usepackage{amssymb}
\usepackage{algorithm}
\usepackage{algorithmic}
\usepackage{subcaption}
\usepackage[inline]{enumitem}
\usepackage{tabularx}
\usepackage{colortbl}
\usepackage{lscape}
\usepackage{placeins}

\newtheorem{theorem}{Theorem}

\def\paperID{*****} % *** Enter the Paper ID here
\def\confName{CVPR}
\def\confYear{2026}

\title{World Action Critic (WAC) Model: Joint Foresight and Criticism for Policy Learning}

\author{Xiangteng Zhang$^{1}$\thanks{Equal contribution.}\quad Haoyuan Liu$^{1}$\footnotemark[1]\quad Yang Guan$^{1}$\quad Jingliang Duan$^{1}$\quad Yao Lyu$^{1}$ \\ 
Tao Zhang$^{3}$\quad Shuai Zhang$^{4}$ \quad Kehua Sheng$^{4}$ \quad Bo Zhang$^{4}$ \quad Shengbo Eben Li$^{1,2}$\thanks{Corresponding author.} \\[0.5em]
$^{1}$School of Vehicle and Mobility, Tsinghua University\quad 
$^{2}$College of AI, Tsinghua University\\
$^{3}$SunRisingAI Lab \quad $^{4}$Didi Voyager Labs, Didi Autonomous Driving
}

\begin{document}
\maketitle

\begin{abstract}
World models have shown considerable promise in sequential decision-making by anticipating the consequences of actions and providing predictive foresight for policy learning.
However, learning to reason about counterfactual actions remains challenging, since predictive objectives may underemphasize information critical for action evaluation, while unexecuted actions lack observed futures for direct supervision.
To address these challenges, we propose the World Action Critic (WAC) Model, which proposes multimodal action candidates and jointly learns candidate-conditioned future prediction and consequence evaluation. 
In our proposed framework, critic supervision directly shapes intermediate predictive representations during rollout, encouraging them to retain information needed to distinguish action quality. 
Observed futures supervise prediction for executed actions, while external task-quality targets provide evaluation signals for counterfactual candidates.
Through joint optimization, critic signals from these candidates propagate into predictive representations without requiring their observed futures.
Our theoretical analysis characterizes critic-guided feature retention and bounds candidate-selection regret. Applied to autonomous driving, WAC achieves 93.4 PDMS on NAVSIM-v1, demonstrating the effectiveness of jointly learning predictive foresight and consequence evaluation.
\end{abstract}

\input{1_intro}
\input{2_related_work}
\input{3_method}
\input{4_experiments}
\input{5_conclusion}
{
    \small
    \bibliographystyle{ieeenat_fullname}
    \bibliography{refs}
}

\appendix
\input{6_appendix}

\end{document}

%% file: 1_intro.tex
\section{Introduction}
\label{sec:introduction}

World models enable agents to anticipate how their environments evolve under different actions, supporting policy learning in reinforcement learning~\citep{zhan2025Bootstrap,wang2025Offpolicy, ha2018World,guan2021Direct, hafner2025Mastering}, robot control~\citep{zhang2026Langevin,zhan2026Harmonized}, and autonomous driving~\citep{gao2022Enhance,li2026WorldEngine, wang2024Driving}. By predicting future states from current observations and actions, they provide a basis for reasoning about the consequences of alternative actions before execution. This capability also supports efforts to build physics-native intelligence, as explored in models such as Phi-WM~\citep{sunrisingai2026ActEffect}. Effective decision making therefore requires predictive representations that preserve the information needed to compare these consequences.

\begin{figure*}[t]
    \centering
    \includegraphics[width=\textwidth]{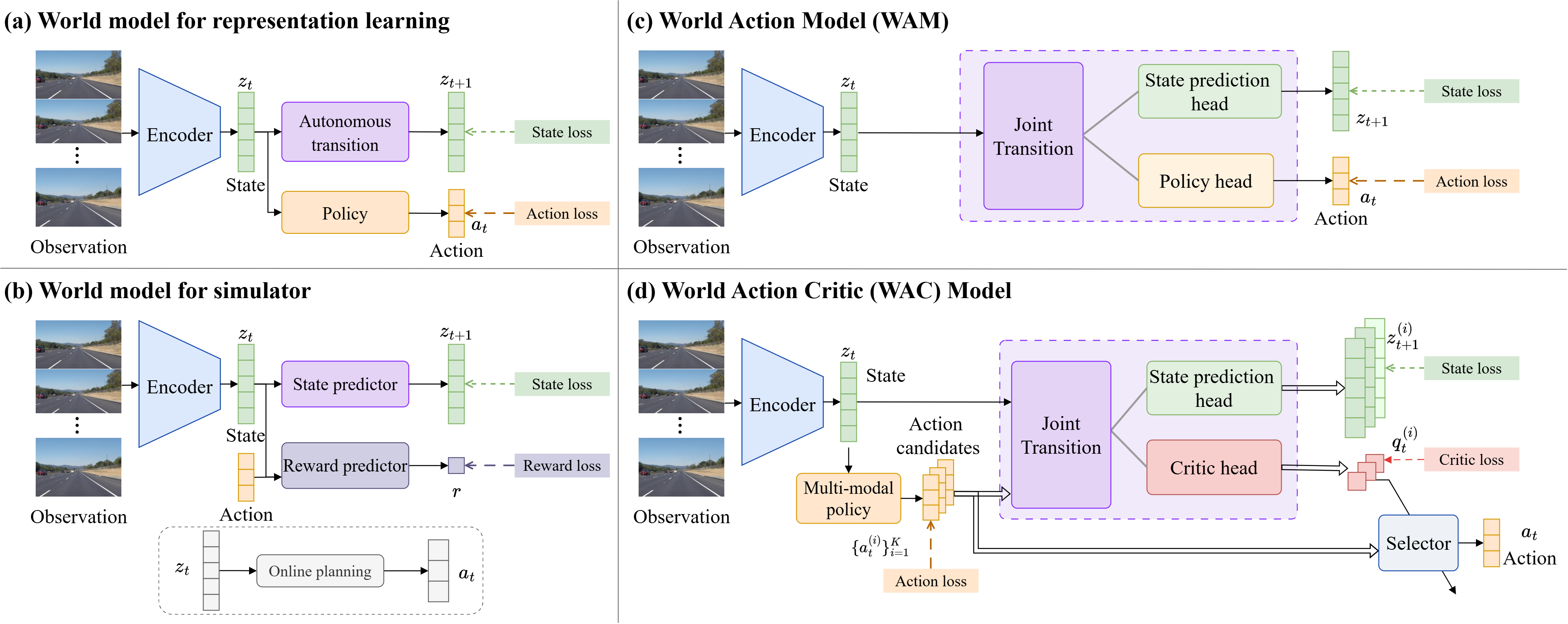}
    \caption{\textbf{Overview of world models for decision making.} (a) Future prediction provides an auxiliary learning signal for policy features. (b) An action-conditioned simulator supplies imagined states to a planner for reward-based evaluation and search. (c) WAM jointly models future states and actions for action generation. (d) WAC jointly learns candidate-conditioned future prediction and consequence evaluation. Factual future targets ground prediction, while candidate-quality supervision shapes predictive representations for counterfactual action comparison.}
    \label{fig:main}
\end{figure*}

Existing approaches integrate world models into policy learning through three main approaches (Figure~\ref{fig:main}). 
First, predictive representation learning uses future-state prediction as an auxiliary objective to enrich the features consumed by a policy~\citep{min2024DriveWorld,li2024Enhancing,bousmalis2023RoboCat, lan2023SEPT}.
Second, simulator-based methods use action-conditioned imagined transitions for reward-based policy optimization or model-based planning~\citep{hafner2019Learning,hafner2020Dream,wu2022DayDreamer,li2024Think2Drive,mu2021ModelBased}. Predicted states are evaluated using rewards or values~\citep{hansen2024TDMPC2}, while the transition model is primarily learned from predictive objectives. Evaluation therefore provides a downstream criterion for imagined consequences, with limited control over the information retained in the predictive representations themselves.
Third, future-conditioned policies and World Action Models (WAMs) bring prediction into action generation, with recent WAMs jointly modeling future states and actions~\citep{du2023UniversalPolicies,zhu2025Unifieda,bi2025Motus,ye2026World,shi2026DriveWAM}. Their joint modeling of future states and actions can capture multimodal behavioral alternatives, while the relative task quality of these alternatives remains implicit. Counterfactual action comparison requires an explicit criterion for evaluating the consequences associated with each candidate.

Learning world models for counterfactual action evaluation faces two challenges. First, predictive representations can be mismatched with the information needed for decision making. Under limited representation capacity, future prediction can favor dominant environmental variation over subtle details that distinguish action quality, making predictive accuracy an insufficient criterion for decision value~\citep{farahmand2017ValueAware,grimm2020ValueEquivalence,voelcker2023Value,gao2022Enhance}. Second, counterfactual actions lack direct future supervision. A recorded transition provides an observed future only for the executed action, while decisions require comparing unexecuted alternatives at the same decision point~\citep{zhang2026Identifiability}. External evaluators can supply task-quality targets through simulation and scoring, but these targets leave complete counterfactual future states unspecified. The challenge is to use this evaluation supervision to learn decision-relevant predictions beyond actions with observed futures.

To address these challenges, we propose the \textbf{World Action Critic (WAC) Model}, which jointly learns candidate-conditioned future prediction and consequence evaluation. A multimodal policy first proposes diverse action candidates. Each candidate conditions a recurrent transition--critic Transformer that predicts future latent states and accumulates evidence for evaluating the candidate over the full horizon. Asymmetric attention allows the critic to read intermediate predictive features while keeping critic tokens hidden from the transition stream. Critic supervision propagates through these features during joint training, encouraging them to preserve information needed to distinguish action quality.

WAC uses complementary supervision to extend this learning to counterfactual actions. Observed futures supervise prediction for executed actions, while an external evaluator supplies candidate-level quality targets for both executed and counterfactual actions. Critic gradients optimize the predictive features along each candidate's rollout, allowing unexecuted alternatives to guide representation learning even when their future observations are unavailable. Building on existing consequence-based trajectory evaluation~\citep{zheng2025World4Drive,li2025EndtoEnd}, WAC couples evaluation to intermediate features throughout recurrent prediction under this complementary supervision. The resulting scores provide a common criterion for comparing multimodal candidates based on their predicted consequences.

Our theoretical analyses address these two learning challenges: they characterize how critic supervision promotes decision-relevant feature retention in a linear setting and bound candidate-selection regret in terms of critic fitting error, evaluator error, and candidate coverage. We instantiate WAC for autonomous driving and evaluate it on NAVSIM~\citep{dauner2024NAVSIM}, achieving 93.4 PDMS on NAVSIM-v1. Controlled ablations examine factual future supervision, recurrent rollout, and critic gradients into predictive features, supporting the complementary roles of future prediction and consequence evaluation.

Our main contributions can be summarized as follows:
\begin{itemize}
    \item We propose WAC, which jointly learns future prediction and consequence evaluation for counterfactual action reasoning, using critic supervision to shape predictive representations for decision making.
    \item We develop a recurrent transition--critic architecture with asymmetric attention and complementary supervision, grounding prediction in factual futures while allowing candidate-quality targets to guide learning from counterfactual actions.
    \item We provide theoretical analyses of critic-guided predictive feature retention and candidate selection, showing how evaluation supervision promotes action-relevant future representations and establishing a regret bound under evaluator accuracy and candidate coverage assumptions.
\end{itemize}

%% file: 2_related_work.tex
\section{Related Work}
\label{sec:related_work}

\subsection{World Models for Planning and Control}

World models provide predictive representations through which agents can reason about actions before execution. Latent dynamics support imagination-based policy optimization, trajectory optimization, and visual planning, as exemplified by Dreamer, TD-MPC2, DINO-WM, and BOOM~\citep{hafner2020Dream,hansen2024TDMPC2,zhou2025DINOWM,zhan2025Bootstrap}. World models also support exploration augmentation by synthesizing transitions for off-policy learning~\citep{wang2025Offpolicy}. This predictive interface extends to physical robot control and autonomous driving, where action-conditioned futures inform behavior learning and planning~\citep{wu2022DayDreamer,wang2024Driving,chen2021Interpretable,guan2022Integrated,guan2026Enhanced}. Across these settings, learned dynamics provide a substrate for decision making: policies or planners use predicted states together with rewards, values, or task costs~\citep{li2023Reinforcement}. Prediction and evaluation often remain identifiable components, even when trained jointly. This leaves an important design question beyond whether imagination improves control: how should an evaluation objective shape the information retained in the predictive representation itself?

\subsection{World Action Models}

World Action Models (WAMs) bring future prediction into action generation, extending action-conditioned dynamics toward joint modeling of environment evolution and behavior. UniPi illustrates an early future-to-action interface by translating generated visual plans into actions through inverse dynamics~\citep{du2023UniversalPolicies}. More recent models, including WorldVLA, Motus, and DriveWAM, couple future and action modeling through shared generative or temporal representations~\citep{cen2025WorldVLA,bi2025Motus,shi2026DriveWAM}. Their common interface uses predictive knowledge to generate or refine behavior, with action preferences potentially encoded in the learned generation distribution. Such preferences do not necessarily provide an explicit task-quality criterion for comparing externally proposed candidates. WAC addresses this complementary problem: evaluating candidate actions supplied by a proposal policy through their predicted consequences, and using evaluation supervision to shape those predictions.

\subsection{World Models for Value Estimation and Candidate Evaluation}

Prediction and evaluation have been connected at several levels. Value-aware model learning constrains predictive dynamics by downstream decision objectives, recognizing that uniform transition accuracy need not preserve the information most useful for control~\citep{farahmand2017ValueAware,grimm2020ValueEquivalence,voelcker2023Value}. At the representation level, World Value Models and World Critic Models connect world modeling with value learning to assess task progress or support policy improvement~\citep{wang2026World,fei2026WCM}. At the candidate level, driving methods use predicted consequences to compare proposed behaviors: World4Drive and WoTE connect future modeling to trajectory selection or scoring~\citep{zheng2025World4Drive,li2025EndtoEnd}. Together, these research directions connect predictive modeling with decision objectives through value learning and consequence-based evaluation. For candidate-conditioned decision making, this relationship raises two linked questions: which predictive features evaluation should preserve, and what candidate labels can support when future observations are missing. Building on these connections, WAC focuses on jointly learning candidate-conditioned foresight and consequence criticism under asymmetric supervision: observed future targets supervise factual transitions, while critic labels supervise a broader set of candidate actions. Our analysis examines how these signals influence feature retention and candidate selection without observed futures for every candidate. A recurrent dual-stream realization lets criticism read intermediate predictive features and, through joint optimization, shape the representations used for future prediction.

%% file: 3_method.tex
\section{World Action Critic (WAC) Model}
\label{sec:method}

In this section, we present the World Action Critic (WAC) Model, which unifies action proposal, future prediction, and consequence evaluation for candidate selection. We formulate their coupling within a joint model and develop a recurrent architecture that learns predictive representations from factual futures and candidate-level quality supervision.

\subsection{From Action Generation to Candidate Evaluation}
\label{sec:modeling}

Let $z=E(o_{\leq t})$ denote the latent representation of the available observations and task context. We use $a$ to denote an action or action sequence and $z'$ to denote the corresponding future latent state or sequence. The coupling between future prediction and action generation in a WAM can be expressed as
\begin{equation}
 p(z',a\mid z)=p(z'\mid z)\,p(a\mid z,z'),
\label{eq:wam_factorization}
\end{equation}
where predictive future context informs action generation.

WAC extends this joint formulation with an explicit critic that evaluates the consequences of proposed actions. With $q$ denoting the task-quality score predicted by the critic, the model factorizes as
\begin{equation}
p(z',a,q\mid z)
=\underbrace{\pi(a\mid z)}_{\text{actor}}\,
\underbrace{p(z'\mid z,a)}_{\text{world}}\underbrace{p(q\mid z,a,z')}_{\text{critic}}.
\label{eq:wacm_factorization}
\end{equation}
The policy $\pi$ proposes a candidate set $\mathcal A_N=\{a^{(i)}\}_{i=1}^{N}$. For each candidate $a^{(i)}$, the world model predicts its corresponding future $\widehat z'^{(i)}$, and the critic produces a quality estimate $\widehat q^{(i)}$ conditioned on the current latent state, the candidate, and its predicted consequences. The parenthesized superscript $(i)$ indexes candidates, so each action is paired with its own future and evaluation. These factors describe the functional roles within a unified WAC, with the critic drawing on predictive features as the candidate-conditioned future is formed.

The resulting scores provide a common criterion for comparing the candidates. WAC selects the action with the highest predicted quality,
\begin{equation}
\widehat a=a^{(i^\star)},
\qquad
i^\star\in\arg\max_{1\leq i\leq N}\widehat q^{(i)}.
\label{eq:wacm_selection_rule}
\end{equation}
\subsection{Joint Prediction and Criticism}
\label{sec:architecture}

\begin{figure}[t]
\centering
\includegraphics[width=\linewidth]{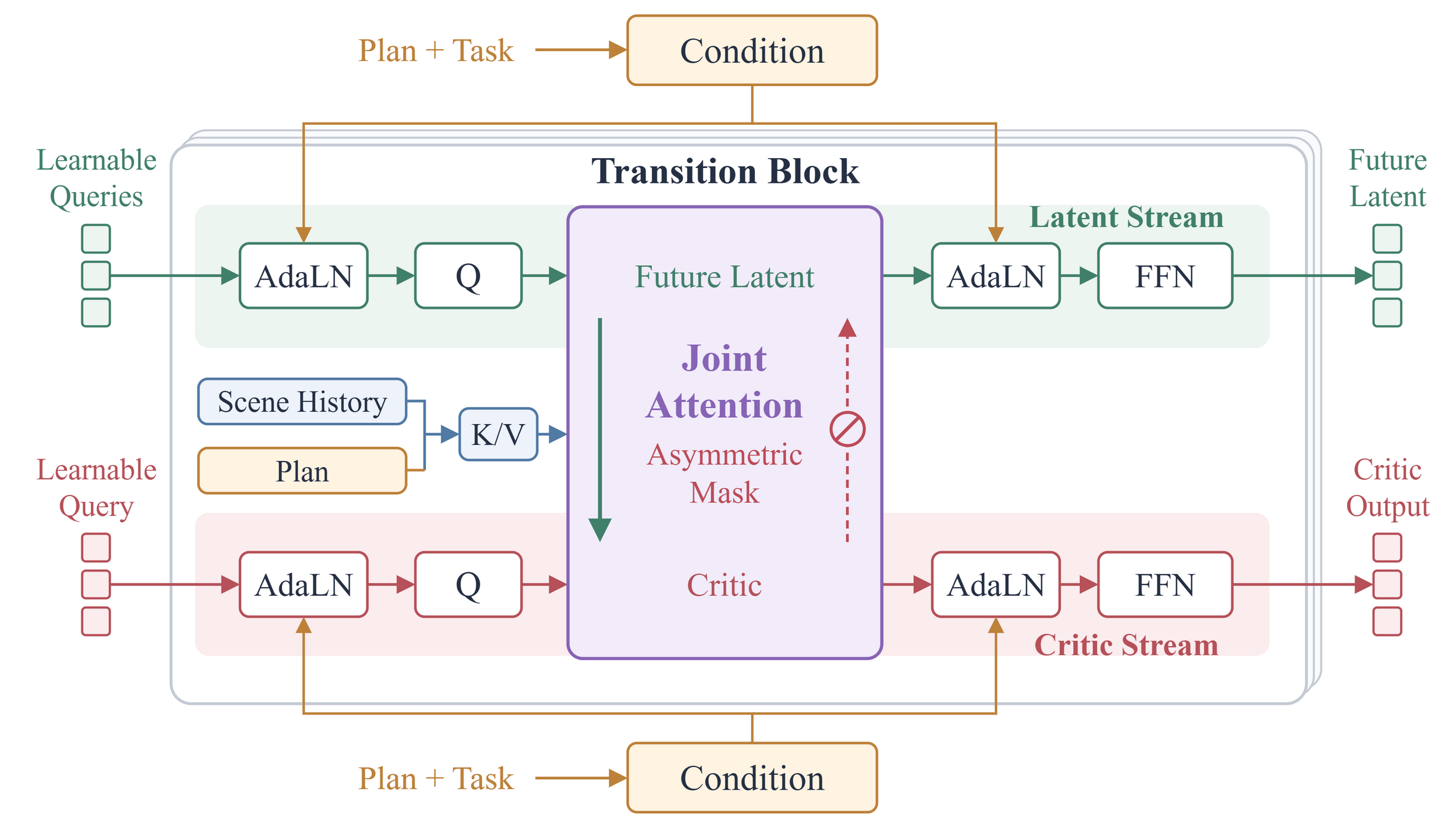}
\caption{\textbf{Joint prediction and criticism in WAC.} Given latent state $z$ and candidate action $a$, WAC jointly predicts future latent state $z'$ and quality score $q$. Asymmetric attention lets the critic read predictive features while keeping critic tokens hidden from latent prediction. Conditions are injected through adaptive layer normalization (AdaLN).}
\label{fig:wacm_architecture}
\end{figure}

The candidate evaluation in Section~\ref{sec:modeling} relies on predicting the consequences of an action from the current latent state. We implement this dependency with a joint prediction--critic module that predicts future latent states and extracts evidence for action evaluation during their generation. As illustrated in Figure~\ref{fig:wacm_architecture}, asymmetric attention connects future prediction to criticism within each layer, while recurrent application extends this interaction over the prediction horizon.

\paragraph{Joint prediction--critic module.}
WAC couples latent prediction and action evaluation through a shared sequence of joint Transformer layers. The action encoder $E_a$ maps the complete candidate action $a$ to a token $E_a(a)$ for attention; this embedding is also combined with the task condition for AdaLN. Starting from the current latent state, the module recurrently predicts action-conditioned futures and accumulates critic evidence. Each candidate $a^{(i)}$ induces a separate prediction under shared model parameters.

We omit the candidate index below and use $k$ for rollout steps. The history memory $M_k$ contains latent states $z_{0:k}$ and critic tokens $\mathbf q_{0:k}$. At each step, latent prediction uses learned queries conditioned on the candidate action and task, while the critic query is initialized from $\mathbf q_k$. The initial critic token $\mathbf q_0$ is learned. A stack of $L$ joint layers updates the prediction and critic tokens, with the same stack and complete candidate action reused across rollout steps.

\paragraph{Asymmetric joint attention.}
Within a joint layer, latent prediction first attends to the latent history together with the action token, followed by a residual feed-forward update. The critic then attends to the updated predictive features, the full history, the action token, and its own current representation. Writing the layer-$\ell$ prediction and critic tokens as $\mathbf z^{\ell}$ and $\mathbf q^{\ell}$, their visible key/value sources are
\begin{equation}
\begin{aligned}
\mathcal S_z^{\ell} &= [M_k^z,E_a(a)],\\
\mathcal S_q^{\ell} &= [M_k,E_a(a),\mathbf z^{\ell+1},\mathbf q^{\ell}],
\end{aligned}
\label{eq:attention_visibility}
\end{equation}
where $M_k^z$ contains the latent states, and bold symbols denote intermediate token representations. An asymmetric mask excludes historical critic tokens from latent prediction. The critic can therefore evaluate emerging future features at every layer, while latent prediction depends on state history and the candidate action. Critic gradients propagate through the predictive features used for evaluation, allowing quality supervision to shape their representations.

Prediction and criticism use separate attention projections and feed-forward networks. Action and task conditions are injected through adaptive layer normalization (AdaLN)~\citep{peebles2023Scalable}, which generates feature-wise scale and shift parameters for the normalized attention queries and feed-forward inputs. Rotary position embeddings encode temporal and spatial coordinates.

After the final layer, normalized prediction tokens produce a residual update to the latent state, while the normalized critic token becomes $\mathbf q_{k+1}$. These outputs are appended to the history for the next step. After $H$ steps, a prediction head maps the final critic token $\mathbf q_H^{(i)}$ to the scalar score $\widehat q^{(i)}$ and auxiliary quality components. Candidates are evaluated in parallel and selected using \eqref{eq:wacm_selection_rule}.

\subsection{Learning from Factual Futures and Candidate Critiques}
\label{sec:joint_training}

We train WAC through complementary supervision for future prediction and critic learning. The recorded action provides an observed future for learning latent dynamics, while an external evaluator supplies critic targets for candidate actions. This difference in supervision allows the model to learn from factual transitions and extend evaluation to a broader set of proposed actions.

\paragraph{Factual future supervision.}
Let $a^{\mathrm f}$ denote the recorded action and $z_k^{\mathrm f}$ its future latent target at step $k=1,\ldots,H$. Conditioned on $a^{\mathrm f}$ and the initial latent state, WAC recurrently predicts $\widehat z_k^{\mathrm f}$. Future observations provide the corresponding latent targets, which are treated as fixed supervision through a stop-gradient operation. The prediction objective is
\begin{equation}
\mathcal L_z = \frac{1}{H}\sum_{k=1}^{H}
\ell_z\!\left(\widehat z_k^{\mathrm f},\operatorname{sg}(z_k^{\mathrm f})\right),
\label{eq:future_loss}
\end{equation}
where $\ell_z$ measures the discrepancy between predicted and target latent states, and $\operatorname{sg}$ stops gradients through the target representation. This objective grounds the learned dynamics in the consequences of the recorded action. The same predictive model is applied to other candidates, whose evaluation is supervised through the critic.

\paragraph{Critic learning.}
For each candidate $a^{(i)}$, let $q^{(i)}$ denote its evaluator-provided target and $\widehat q^{(i)}$ the critic prediction. During training, we include the recorded action as $a^{(N+1)}=a^{\mathrm f}$ alongside the $N$ proposed candidates. The critic objective is
\begin{equation}
\mathcal L_q=\frac{1}{N+1}
\sum_{i=1}^{N+1}
\ell_q\!\left(\widehat q^{(i)},q^{(i)}\right),
\label{eq:critic_loss}
\end{equation}
where $\ell_q$ measures the discrepancy between the critic prediction and its supervision target. Candidate-level targets therefore extend learning beyond the recorded action, even when the corresponding future latent targets are unavailable.

\paragraph{Joint optimization.}
The proposal policy $\pi$ is supervised by the recorded action $a^{\mathrm f}$ using a best-of-$N$ regression objective. Among the $N$ generated candidates, we select the one closest to the ground truth and apply the action loss to that candidate:
\begin{equation}
i_\pi\in\arg\min_{1\leq i\leq N}\ell_a\!\left(a^{(i)},a^{\mathrm f}\right),
\qquad
\mathcal L_\pi=\ell_a\!\left(a^{(i_\pi)},a^{\mathrm f}\right),
\label{eq:proposal_loss}
\end{equation}
where $\ell_a$ is the distance over the action sequence. This objective encourages the proposal policy to generate a candidate that matches the demonstrated action. Together with future prediction and critic learning, the joint objective is
\begin{equation}
\mathcal L=\lambda_\pi\mathcal L_\pi
+\lambda_z\mathcal L_z
+\lambda_q\mathcal L_q,
\label{eq:wacm_total_objective}
\end{equation}
where the nonnegative weights balance the three objectives. The losses above are defined for a training instance and averaged over the training data. The choices of $\ell_z$ and $\ell_q$ depend on the prediction targets and critic supervision of the task.

The joint architecture connects these learning signals through the predictive features used by the critic. Factual future supervision constrains latent dynamics, while critic supervision propagates through asymmetric attention and shapes the predictive representations used to compare candidates. Each evaluated candidate thus contributes a learning signal to its predicted future representation. At inference, the proposal policy supplies candidate actions, and WAC predicts their consequences and selects an action using its learned critic according to \eqref{eq:wacm_selection_rule}.
\subsection{Why Joint Prediction and Criticism Help}
\label{sec:theory}

Our analysis addresses the mismatch between predictive representation learning and decision-oriented evaluation, and incomplete counterfactual future supervision for candidate actions. We first show how critic supervision reweights predictive representations toward decision-relevant future information. We then establish how candidate-level critic supervision supports action selection under coverage of the proposed actions, including those with unobserved futures. The two results explain the roles of criticism in representation learning and decision making. Full assumptions and proofs are provided in Appendix~\ref{sec:proofs}.
\paragraph{Learning predictive representations for criticism.}
Future prediction naturally prioritizes features that explain the variation of future states, although these features need not be the most informative for evaluating actions. Joint learning of prediction and criticism introduces an additional learning signal that can promote future features according to their relevance to action evaluation. We formalize this effect in a simple linear setting. Let the centered future latent representation be $z'\in\mathbb R^D$ with covariance $\Sigma\succ0$, and let the critic target depend on the future representation as
\begin{equation}
q=b^\top z'+\xi,
\end{equation}
where $\mathbb E[\xi\mid z']=0$ and $\mathbb E[\xi^2]<\infty$. We whiten the future representation as $\widetilde z=\Sigma^{-1/2}z'$ and consider a $d$-dimensional predictive representation $U^\top\widetilde z$, where $1\leq d<D$ and $U^\top U=I_d$. Prediction and criticism are modeled by optimal linear heads with squared losses, and their relative weight is controlled by $\lambda\geq0$. Define the critic-relevant future direction as
\begin{equation}
v=\Sigma^{1/2}b.
\end{equation}

\begin{theorem}[Critic-guided predictive importance]
\label{thm:feature_retention}
Under the setting above, joint learning induces the importance matrix
\begin{equation}
K_\lambda
=
\underbrace{\Sigma}_{\text{predictive importance}}
+
\underbrace{\lambda vv^\top}_{\text{decision importance}}.
\label{eq:critic_guided_importance}
\end{equation}
An optimal $d$-dimensional predictive subspace is given by
\begin{equation}
\operatorname{span}(U_\lambda)
=
\operatorname{TopEig}_d(K_\lambda).
\label{eq:critic_guided_subspace}
\end{equation}
where $\operatorname{TopEig}_d$ denotes a subspace spanned by eigenvectors associated with the $d$ largest eigenvalues; ties admit any maximizing choice.
\end{theorem}

Theorem~\ref{thm:feature_retention} shows that the prediction objective favors future directions with high predictive importance, while critic supervision additionally promotes directions that are informative for action evaluation. Consequently, a decision-relevant future feature can be retained even when it is not dominant under prediction alone. In WAC, this mechanism is realized by allowing gradients from $\mathcal L_q$ to propagate through the predictive features consumed by the critic. The complete derivation and a two-dimensional example are provided in Appendix~\ref{sec:proof_retention}.

\paragraph{Action selection under candidate supervision.}
The second issue arises from the asymmetric supervision available during training. The recorded action provides an observed future latent target, while proposed candidate actions generally have no corresponding counterfactual future observations. Their external critic targets can nevertheless provide direct supervision for candidate comparison. Let $\mu$ denote the joint distribution of critic-supervised $(z,a)$ pairs. Let $\nu$ denote the distribution obtained by sampling a test latent state and its proposed candidate set and then choosing one of the $N$ candidates uniformly. We write $q^\star(z,a)$ for the intended task score, $q_{\mathrm{ext}}(z,a)=\mathbb E[q\mid z,a]$ for the evaluator's conditional mean, and $\widehat q(z,a)$ for WAC's predicted score. Define
\begin{align}
\epsilon_{\mathrm{fit}}&=\mathbb E_\mu\left(\widehat q-q_{\mathrm{ext}}\right)^2,\\
\epsilon_{\mathrm{ext}}&=\mathbb E_\mu\left(q_{\mathrm{ext}}-q^\star\right)^2,
\label{eq:critic_error_terms}
\end{align}
where $\epsilon_{\mathrm{fit}}$ measures critic fitting error and $\epsilon_{\mathrm{ext}}$ measures the discrepancy between the external evaluator and the intended task score.

\begin{theorem}[Candidate selection under critic supervision]
\label{thm:candidate_selection}
Assume $\nu\ll\mu$ with $d\nu/d\mu\leq\kappa<\infty$ $\mu$-almost surely, and that the errors in \eqref{eq:critic_error_terms} are finite. Let $a^\star$ maximize $q^\star$ over the proposed candidate set $\mathcal A_N$, and let $\widehat a$ be selected by \eqref{eq:wacm_selection_rule}. Then
\begin{equation}
\begin{aligned}
&\mathbb E\left[q^\star(z,a^\star)-q^\star(z,\widehat a)\right]\\
&\qquad\leq
\sqrt{2N\kappa}
\left(
\sqrt{\epsilon_{\mathrm{fit}}}
+
\sqrt{\epsilon_{\mathrm{ext}}}
\right).
\end{aligned}
\label{eq:candidate_regret}
\end{equation}
The expectation includes the joint distribution of the proposed candidates and allows dependence among them.
\end{theorem}

The bound shows that candidate selection depends on the accuracy of WAC's critic predictions against the evaluator's targets, the alignment of that evaluator with the intended task objective, and the coverage of test candidates by the critic-supervised distribution. Candidate-level critic targets can therefore support reliable comparison without requiring counterfactual future observations for every proposed action. Together, Theorems~\ref{thm:feature_retention} and~\ref{thm:candidate_selection} show how joint criticism can influence both the learned future representation and the subsequent selection among candidate actions.

%% file: 4_experiments.tex
\section{Experiments}
\label{sec:experiments}

\subsection{Dataset and Evaluation Metrics}
\label{sec:dataset_metrics}

\paragraph{Dataset.}
We conduct our experiments on the NAVSIM benchmark, which is built upon OpenScene and ultimately derived from the nuPlan dataset. It covers diverse urban driving scenarios. Following prior work, we train on the \texttt{navtrain} split and evaluate on the complete \texttt{navtest} split.

\paragraph{Evaluation Metrics.}
We evaluate planning on NAVSIM-v1 with the PDM Score (PDMS). It consists of five sub-metrics: No at-fault Collisions (NC), Drivable Area Compliance (DAC), Time to Collision (TTC), Ego Progress (EP), and Comfort (C). NC and DAC serve as multiplicative penalties, while TTC, EP, and C are combined as a weighted average. The overall PDMS is defined as:
\begin{equation}
\mathrm{PDMS}
=
\mathrm{NC}
\cdot
\mathrm{DAC}
\cdot
\frac{
5\mathrm{EP}
+
5\mathrm{TTC}
+
2\mathrm{C}
}{12}.
\label{eq:pdms}
\end{equation}

\iffalse
For NAVSIM-v2, we additionally report the Extended PDM Score (EPDMS), which extends PDMS with Driving Direction Compliance (DDC), Traffic Light Compliance (TLC), Lane Keeping (LK), History Comfort (HC), and Extended Comfort (EC). Specifically, the EPDMS is defined as
\begin{equation}
\mathrm{EPDMS}
=
\mathrm{NC}
\cdot
\mathrm{DAC}
\cdot
\mathrm{DDC}
\cdot
\mathrm{TLC}
\cdot
\frac{
5\mathrm{TTC}
+
5\mathrm{EP}
+
2\mathrm{LK}
+
2\mathrm{HC}
+
2\mathrm{EC}
}{16}.
\label{eq:epdms}
\end{equation}
\fi

% Queue the wide table early to place it beside the main-results discussion.
\begin{table*}[!t]
\centering
\caption{Comparison with end-to-end, vision-language-action, and world-model/world-action planners on NAVSIM-v1. We report overall PDMS and its safety, road-compliance, comfort, and progress components; higher is better for all metrics. Baseline results follow their reported sensor and training configurations. Using only front-camera observations and ego status, WAC achieves the highest overall PDMS and ego progress among the compared methods.}
\label{tab:pdms}
\small
\setlength{\tabcolsep}{4pt}
\renewcommand{\arraystretch}{0.90}
\begin{tabular*}{\textwidth}{@{\extracolsep{\fill}}lcccccc}
\toprule
Method & NC$\uparrow$ & DAC$\uparrow$ & TTC$\uparrow$ & Comf.$\uparrow$ & EP$\uparrow$ & PDMS$\uparrow$ \\
\midrule
\rowcolor{gray!12}\multicolumn{7}{c}{\textit{End-to-End Planning}} \\
DiffusionDriveV2~\citep{zou2025DiffusionDriveV2} & 98.3 & 97.9 & 94.8 & 99.9 & 87.5 & 91.2 \\
iPad (ViT-L)~\citep{guo2025iPad} & 99.2 & 97.4 & 96.3 & 99.7 & 87.8 & 91.7 \\
SparseDriveV2~\citep{sun2026SparseDriveV2} & 98.5 & 98.4 & 95.0 & 99.9 & 88.6 & 92.0 \\
NaviHydra (V4096)~\citep{wu2025NaviHydra} & 98.7 & 98.6 & 96.2 & 100.0 & 88.7 & 92.7 \\
DrivoR (train)~\citep{kirby2026DrivoR} & 98.9 & 98.3 & 96.2 & 100.0 & 89.1 & 93.1 \\
Drive-JEPA (ViT-L, v1)~\citep{wang2026DriveJEPA} & 99.1 & 98.2 & 95.9 & 99.9 & 90.8 & 93.3 \\
\addlinespace[2pt]
\rowcolor{gray!12}\multicolumn{7}{c}{\textit{Vision-Language-Action Planning}} \\
AutoVLA~\citep{zhou2025AutoVLA} & 98.4 & 95.6 & 98.0 & 99.9 & 81.9 & 89.1 \\
DriveVLA-W0 (Anchor)~\citep{li2025DriveVLAW0} & 98.7 & 99.1 & 95.3 & 99.3 & 83.3 & 90.2 \\
ReCogDrive~\citep{li2025ReCogDrive} & 97.9 & 97.3 & 94.9 & 100.0 & 87.3 & 90.8 \\
DriveWorld-VLA~\citep{jia2026DriveWorldVLA} & 99.1 & 98.2 & 96.1 & 100.0 & 85.9 & 91.3 \\
DynVLA~\citep{shang2026DynVLA} & 98.6 & 98.7 & 95.5 & 100.0 & 86.8 & 91.7 \\
HybridDriveVLA~\citep{ang2026HybridDriveVLA} & 98.6 & 98.6 & 96.2 & 100.0 & 87.3 & 92.1 \\
LatentVLA (iPad, v1)~\citep{xie2026LatentVLA} & 98.9 & 98.2 & 95.2 & 100.0 & 88.2 & 92.4 \\
WCog-VLA (2B)~\citep{yan2026WCogVLA} & 99.4 & 98.8 & 98.5 & 100.0 & 87.1 & 92.9 \\
\addlinespace[2pt]
\rowcolor{gray!12}\multicolumn{7}{c}{\textit{World-Model and World-Action Planning}} \\
Epona~\citep{zhang2025Epona} & 97.9 & 95.1 & 93.8 & 99.9 & 80.4 & 86.2 \\
WorldRFT~\citep{yang2026WorldRFT} & 97.8 & 96.8 & 94.0 & 100.0 & 81.7 & 87.8 \\
WoTE~\citep{li2025EndtoEnd} & 98.5 & 96.8 & 94.9 & 99.9 & 81.9 & 88.3 \\
DriveLaW~\citep{xia2026DriveLaW} & 99.0 & 97.1 & 96.7 & 100.0 & 81.3 & 89.1 \\
DriveDreamer-Policy~\citep{zhou2026DriveDreamerPolicy} & 98.4 & 97.1 & 95.1 & 100.0 & 83.5 & 89.2 \\
DriveWAM~\citep{shi2026DriveWAM} & 98.3 & 98.1 & 95.2 & 100.0 & 84.3 & 90.1 \\
ReWorld~\citep{xia2026ReWorld} & 99.1 & 98.2 & 97.7 & 99.8 & 82.0 & 90.4 \\
DriveFuture~\citep{hong2026DriveFuture} & 98.8 & 99.1 & 95.4 & 100.0 & 84.2 & 90.7 \\
Auto-JEPA~\citep{yang2026AutoJEPA} & 98.4 & 98.3 & 95.0 & 100.0 & 87.1 & 91.3 \\
SimWAM~\citep{zhao2026SimWAM} & 98.4 & 98.7 & 95.5 & 100.0 & 86.4 & 91.5 \\
\midrule
\textbf{WAC (Ours)} & 99.0 & 98.3 & 95.9 & 100.0 & 91.0 & \textbf{93.4} \\
\bottomrule
\end{tabular*}
\end{table*}

\subsection{Implementation Details}
\label{sec:implementation_details}

Our visual backbone is based on V-JEPA 2~\citep{assran2025VJEPA}. We initialize the ViT-L encoder from the driving-video-pretrained Drive-JEPA checkpoint~\citep{wang2026DriveJEPA} and jointly fine-tune it with the planning modules using RADAR~\citep{ren2026momentum}, with mixed precision, and constant learning rates. We summarize the architecture and training hyperparameters in Appendix~\ref{sec:implementation_appendix}.

Each input consists of two consecutive front-view observations. The model proposes 32 candidate trajectories, each containing eight future poses over 4s, and evaluates them in parallel with shared WAC parameters. Each recurrent prediction step advances by four poses, yielding two prediction steps over the planning horizon. At each step, the complete candidate trajectory and driving command condition future-state prediction and critic updates.

In the driving instantiation, the state $z_k$ in Section~\ref{sec:architecture} comprises scene latents and ego dynamics, both retained in the history memory. The future-prediction objective $\mathcal L_z$ combines their respective Smooth L1 losses against factual targets, while NAVSIM PDM evaluations supervise six critic outputs: NC, DAC, EP, TTC, comfort, and aggregate PDMS. At inference, we select the candidate with the highest predicted aggregate score. The focal-weighted variant is defined in Appendix~\ref{sec:implementation_appendix}. The ablations use the same NAVSIM-v1 evaluation split and report PDMS in points.

\subsection{Main Results}
\label{sec:main_results}

Table~\ref{tab:pdms} compares WAC with end-to-end, vision-language-action (VLA), and world-model/world-action planners on NAVSIM-v1. Our front-camera planner achieves $93.4$ PDMS, the highest overall score among the compared methods, including DrivoR~\citep{kirby2026DrivoR} and the full Drive-JEPA planner~\citep{wang2026DriveJEPA}. The results show that WAC achieves the highest ego progress while maintaining competitive safety and compliance scores, indicating a favorable balance between planning efficiency, safety, and compliance. The following scoring analysis and controlled ablations examine how this behavior relates to candidate evaluation and joint predictive learning.

\subsection{Candidate Scoring Quality}
\label{sec:scoring_accuracy}

Candidate selection depends on the relative quality assigned to alternative trajectories. We therefore evaluate every proposed candidate with the NAVSIM-v1 evaluator and examine two aspects of the learned scores: whether they preserve the ordering of candidate quality, measured by Spearman correlation and pairwise ranking accuracy, and whether they overrate poor alternatives, measured by low-score overestimation. The analysis covers the complete test split and uses each planner's own candidate pool. Metric definitions are given in Appendix~\ref{sec:scoring_protocol}.

\begin{table}[!tbp]
\centering
\caption{Candidate scoring quality on NAVSIM-v1. PA denotes pairwise ranking accuracy.}
\label{tab:scoring_accuracy}
\small
\setlength{\tabcolsep}{4pt}
\renewcommand{\arraystretch}{1.05}
\begin{tabular*}{\columnwidth}{@{\extracolsep{\fill}}lccc@{}}
\toprule
Method & Spearman$\uparrow$ & PA (\%)$\uparrow$ & \shortstack{Low-score\\overestimation$\downarrow$} \\
\midrule
Drive-JEPA & 0.468 & 74.02 & 44.16 \\
WAC & \textbf{0.551} & \textbf{77.11} & \textbf{34.02} \\
\bottomrule
\end{tabular*}
\end{table}

Table~\ref{tab:scoring_accuracy} shows that WAC improves candidate ordering and reduces overestimation of low-quality alternatives. Consistent gains in Spearman correlation and pairwise ranking accuracy indicate that its scores more reliably distinguish preferable candidates within a scene. Reduced low-score overestimation addresses inflated quality estimates that can promote poor trajectories during selection. The following ablations examine the contributions of joint prediction and evaluation.

% Queue the trajectory figure early so it can share a page with the ablations.
\begin{figure*}[!t]
\centering
\includegraphics[width=0.29\textwidth]{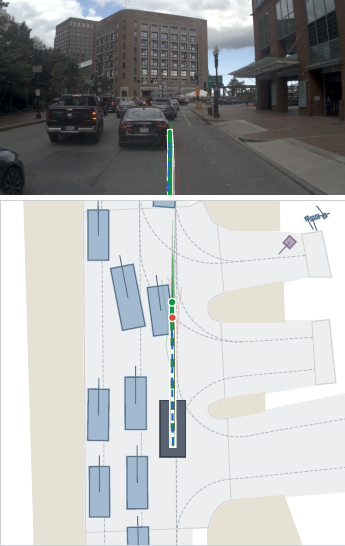}\hfill
\includegraphics[width=0.29\textwidth]{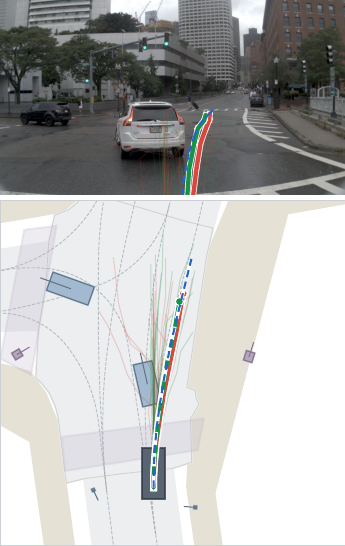}\hfill
\includegraphics[width=0.29\textwidth]{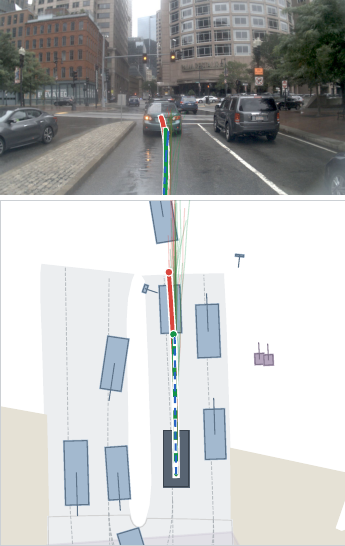}

\includegraphics[width=0.88\textwidth]{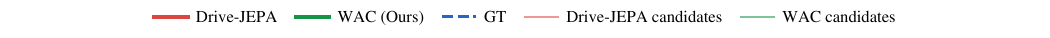}
\caption{Selected trajectories in urban car-following, a signalized urban junction, and a wet urban junction (left to right). Each column pairs the front-camera trajectory projection with its BEV. Red denotes Drive-JEPA, green WAC (Ours), and blue dashed lines the recorded GT. Thin lines show the 32 candidates per method; thick lines show their selections. Endpoint dots in BEV mark four-second predictions. }
\label{fig:wacm_qualitative}
\end{figure*}

\subsection{Ablation Studies}
\label{sec:ablation}

\paragraph{Model structure and supervision.}
Table~\ref{tab:wacm_component_ablation} compares WAC variants trained without focal weighting and with the same four-pose prediction interval. This isolates the effect of recurrent prediction, factual future supervision, and critic-to-transition gradients within the same planner.

\begin{table}[!tbp]
\centering
\small
\caption{WAC structure and supervision ablations on NAVSIM-v1.}
\label{tab:wacm_component_ablation}
\setlength{\tabcolsep}{2pt}
\renewcommand{\arraystretch}{1.05}
\begin{tabular*}{\columnwidth}{@{\extracolsep{\fill}}lcccc@{}}
\toprule
Variant & AR & Future & Grad & PDMS$\uparrow$ \\
\midrule
Detached critic gradients & $\checkmark$ & $\checkmark$ & $\times$ & 92.3 \\
One-shot future prediction & $\times$ & $\checkmark$ & $\checkmark$ & 92.6 \\
No future supervision & $\checkmark$ & $\times$ & $\checkmark$ & 93.0 \\
Full WAC& $\checkmark$ & $\checkmark$ & $\checkmark$ & \textbf{93.3} \\
\bottomrule
\end{tabular*}
\end{table}

Autoregressive prediction outperforms one-shot future prediction. The one-shot variant predicts future latent states in parallel from the initial context, without feeding an imagined state into the next prediction. WAC instead propagates its predicted latent and critic states between steps, allowing later consequence estimates to depend on earlier imagined states. The comparison supports maintaining this sequential dependence when evaluating a candidate over the planning horizon.

Removing factual future-latent and ego-state supervision weakens performance, indicating that future targets contribute beyond trajectory-quality labels alone. Blocking critic gradients from the predictive features causes the largest degradation among these variants. The critic can still learn to score candidates in that setting, but its quality supervision cannot adapt the transition features toward action evaluation. This comparison supports the central joint-learning mechanism: predictive features benefit from being optimized for the consequences that matter to candidate selection. Factual future supervision and critic feedback thus play complementary roles, grounding prediction in observed dynamics and making the learned features useful for decisions.

% Queue the scoring figure before its discussion.
\begin{figure*}[!t]
\centering
\captionsetup[subfigure]{font=small,position=top,justification=centering,skip=4pt}
\begin{subfigure}[t]{0.30\textwidth}
\centering
\caption{Scoring and selection}
\label{fig:scoring_case_joint}
\includegraphics[width=\linewidth]{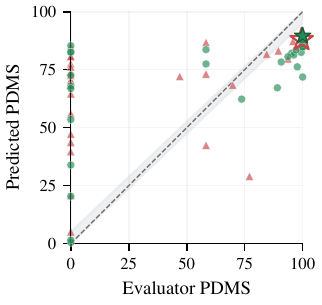}
\end{subfigure}\hfill
\begin{subfigure}[t]{0.30\textwidth}
\centering
\caption{Low-score overestimation}
\label{fig:scoring_case_low}
\includegraphics[width=\linewidth]{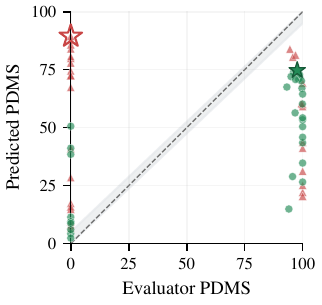}
\end{subfigure}\hfill
\begin{subfigure}[t]{0.30\textwidth}
\centering
\caption{Score agreement}
\label{fig:scoring_case_typical}
\includegraphics[width=\linewidth]{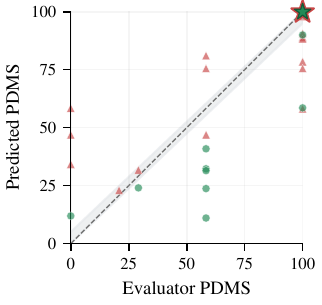}
\end{subfigure}

\includegraphics[width=0.42\textwidth]{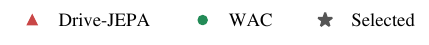}
\caption{Candidate scoring in three selected improvement examples. Each subfigure overlays all 32 Drive-JEPA candidates (red triangles) and all 32 WAC candidates (green circles). Stars mark selections: red outlines for Drive-JEPA and green fills for WAC. The dashed line indicates equality and the shaded band is $\pm5$ PDMS points. Coincident candidates overlap. Each model generates its own candidates.}
\label{fig:scoring_cases}
\end{figure*}

\paragraph{Prediction interval and focal weighting.}
Table~\ref{tab:wacm_step_ablation} varies the prediction interval while keeping the planning horizon fixed. With focal weighting disabled, predicting four poses per step over two rollout steps achieves 93.3 PDMS, compared with 92.8 for two poses per step over four rollout steps. The final model uses the four-pose prediction interval.

\begin{table}[!tbp]
\centering
\small
\caption{Prediction interval and focal-loss ablation on NAVSIM-v1. Every trajectory has eight future poses.}
\label{tab:wacm_step_ablation}
\renewcommand{\arraystretch}{1.05}
\begin{tabular*}{\columnwidth}{@{\extracolsep{\fill}}cccc@{}}
\toprule
Poses per step & Rollout steps & Focal & PDMS$\uparrow$ \\
\midrule
2 & 4 & $\times$ & 92.8 \\
4 & 2 & $\times$ & 93.3 \\
4 & 2 & $\checkmark$ & \textbf{93.4} \\
\bottomrule
\end{tabular*}
\end{table}

\begin{samepage}
With the architecture and rollout length fixed, focal weighting further increases PDMS from 93.3 to 93.4. It emphasizes difficult component predictions, while progress and aggregate PDMS retain ordinary binary cross-entropy.
\end{samepage}

\subsection{Qualitative Analysis}
\label{sec:qualitative}

Figure~\ref{fig:wacm_qualitative} compares WAC with the Drive-JEPA baseline in urban driving scenes. The candidate sets and selected paths show how the planners resolve alternatives, while the NAVSIM evaluator assesses the consequences of their choices.

The car-following example separates immediate collision avoidance from temporal safety: both selections pass the collision check, but only WAC passes TTC. At the signalized junction, the baseline selection violates drivable-area compliance, whereas WAC preserves compliance while maintaining progress. The wet-junction example illustrates a similar distinction in road compliance. These cases highlight improvements in temporal safety and drivable-area compliance. Appendix~\ref{sec:additional_cases} connects further planning examples with their candidate rankings.

\paragraph{Candidate-level scoring and selection.}
Figure~\ref{fig:scoring_cases} connects candidate-level scoring with the final selection. In (a), better agreement with evaluator scores accompanies lower selection regret. In (b), the baseline promotes a poor candidate to its highest predicted score, whereas WAC selects a high-quality alternative. In (c), WAC's candidate scores show closer agreement with the evaluator.

%% file: 5_conclusion.tex
\section{Conclusion}
\label{sec:conclusion}

We presented the World Action Critic (WAC) Model, which combines candidate
proposal, action-conditioned future prediction, and explicit evaluation in
one decision framework. Its recurrent transition and critic streams extract
evaluation evidence as each candidate's future is formed. Directed attention
preserves a prediction-to-criticism dependency, while joint optimization lets
candidate-quality supervision shape the transition features. Observed futures
supervise logged actions, and external critic labels supervise additional
candidates. Our analysis explains decision-relevant feature retention in a
restricted capacity model and bounds candidate-selection error under explicit
coverage and evaluator assumptions. NAVSIM experiments demonstrate the
framework's application to autonomous driving. Extending this empirical
validation beyond driving remains future work.

%% file: 6_appendix.tex
\newpage
\section{Theoretical Analysis}
\label{sec:proofs}

\subsection{Critic-Guided Predictive Representations}
\label{sec:proof_retention}

Let $z'\in\mathbb R^D$ satisfy $\mathbb E[z']=0$ and $\operatorname{Cov}(z')=\Sigma\succ0$. The critic target is $q=b^\top z'+\xi$, where $\mathbb E[\xi\mid z']=0$ and $\sigma_\xi^2=\mathbb E[\xi^2]<\infty$. With $\widetilde z=\Sigma^{-1/2}z'$ and $v=\Sigma^{1/2}b$, we have $\operatorname{Cov}(\widetilde z)=I$ and $q=v^\top\widetilde z+\xi$.

For $1\leq d<D$, consider the representation $U^\top\widetilde z$ with $U^\top U=I_d$. Its optimal linear prediction and critic objectives are
\begin{align}
\mathcal R_z(U)&=\min_A\mathbb E\|z'-AU^\top\widetilde z\|_2^2,
\label{eq:appendix_prediction_risk}\\
\mathcal R_q(U)&=\min_w\mathbb E(q-w^\top U^\top\widetilde z)^2,
\label{eq:appendix_critic_risk}
\end{align}
where $A\in\mathbb R^{D\times d}$ and $w\in\mathbb R^d$. Joint learning minimizes $\mathcal R_z(U)+\lambda\mathcal R_q(U)$ for $\lambda\geq0$.

\begin{proof}[Proof of Theorem~\ref{thm:feature_retention}]
Since $\widetilde z$ has identity covariance and the noise is orthogonal to its linear functions, least squares gives the optimal heads $A=\Sigma^{1/2}U$ and $w=U^\top v$. Substituting them yields
\begin{align}
\mathcal R_z(U)&=\operatorname{tr}(\Sigma)-\operatorname{tr}(U^\top\Sigma U),
\label{eq:optimal_linear_prediction}\\
\mathcal R_q(U)&=\sigma_\xi^2+\|v\|_2^2-\|U^\top v\|_2^2.
\label{eq:optimal_linear_critic}
\end{align}
Consequently,
\begin{equation}
\begin{aligned}
\mathcal R_z(U)+\lambda\mathcal R_q(U)
&=\operatorname{tr}(\Sigma)+\lambda(\sigma_\xi^2+\|v\|_2^2)\\
&\quad-\operatorname{tr}(U^\top K_\lambda U),
\end{aligned}
\label{eq:joint_linear_objective}
\end{equation}
where $K_\lambda=\Sigma+\lambda vv^\top$. The first two terms are independent of $U$, so joint learning maximizes $\operatorname{tr}(U^\top K_\lambda U)$.

Let $\eta_1\geq\cdots\geq\eta_D$ and $r_1,\ldots,r_D$ be the eigenvalues and orthonormal eigenvectors of $K_\lambda$. Then
\begin{equation}
\operatorname{tr}(U^\top K_\lambda U)
=\sum_{j=1}^D\eta_j\|U^\top r_j\|_2^2
\leq\sum_{j=1}^d\eta_j,
\end{equation}
because the weights $\|U^\top r_j\|_2^2$ lie in $[0,1]$ and sum to $d$. Equality is attained by a leading $d$-dimensional eigenspace, with any maximizing choice admissible at a tied eigenvalue. This proves \eqref{eq:critic_guided_subspace}.
\end{proof}

\paragraph{Example of retaining a low-variance feature.}
Take $D=2$, $d=1$, $\Sigma=\operatorname{diag}(1,4)$, and $q=2z'_1$ with zero noise. Prediction alone retains the higher-variance feature $z'_2$, while the critic target depends entirely on $z'_1$. Here $v=(2,0)^\top$, so $K_\lambda=\operatorname{diag}(1+4\lambda,4)$. For $\lambda>3/4$, the leading direction switches to $z'_1$. The critic term thus promotes a low-variance future feature according to its relevance to action evaluation.

Whitening provides a parameterization of rank-$d$ linear representations, with changes of basis absorbed into the output heads.

\subsection{Candidate Selection under Critic Supervision}
\label{sec:proof_selection}

\begin{proof}[Proof of Theorem~\ref{thm:candidate_selection}]
Fix the learned proposal policy and critic, and write $e(z,a)=\widehat q(z,a)-q^\star(z,a)$. For a realized candidate set, let $i^\star$ and $\widehat i$ index the task-optimal and selected candidates. Substituting $q^\star=\widehat q-e$ into the regret gives
\begin{equation}
\begin{aligned}
R&:=q^\star(z,a^{(i^\star)})-q^\star(z,a^{(\widehat i)})\\
&=\widehat q(z,a^{(i^\star)})-\widehat q(z,a^{(\widehat i)})\\
&\quad+e(z,a^{(\widehat i)})-e(z,a^{(i^\star)})\\
&\leq e(z,a^{(\widehat i)})-e(z,a^{(i^\star)}).
\end{aligned}
\end{equation}
The last inequality follows because $\widehat i$ maximizes the predicted score, making the difference of predicted scores nonpositive. Thus regret is controlled by the errors on the selected and task-optimal candidates.

If the indices agree, $R=0$. Otherwise, using $R\geq0$ and $(x-y)^2\leq2(x^2+y^2)$ yields
\begin{equation}
\begin{aligned}
R^2&\leq2\bigl[e(z,a^{(\widehat i)})^2+e(z,a^{(i^\star)})^2\bigr]\\
&\leq2\sum_{i=1}^N e(z,a^{(i)})^2.
\end{aligned}
\end{equation}
To average this bound over test instances, recall that $\nu$ samples one candidate uniformly from each proposed set. Hence
\begin{equation}
\mathbb E_\nu e^2
=\mathbb E\!\left[\frac1N\sum_{i=1}^N e(z,a^{(i)})^2\right].
\end{equation}
This identity holds for any dependence among the candidates. Coverage then transfers the error to the critic-supervised distribution,
\begin{equation}
\mathbb E_\nu e^2
=\mathbb E_\mu\!\left[\frac{d\nu}{d\mu}e^2\right]
\leq\kappa\mathbb E_\mu e^2.
\end{equation}
Combining these steps and applying Cauchy--Schwarz inequality, we have
\begin{equation}
\mathbb E R\leq\sqrt{\mathbb E R^2}
\leq\sqrt{2N\mathbb E_\nu e^2}
\leq\sqrt{2N\kappa}\,\|e\|_{L_2(\mu)}.
\end{equation}
Finally, decompose $e=(\widehat q-q_{\mathrm{ext}})+(q_{\mathrm{ext}}-q^\star)$ into critic fitting error and evaluator discrepancy. The triangle inequality in $L_2(\mu)$ gives
\begin{equation}
\begin{aligned}
\|e\|_{L_2(\mu)}
&\leq\|\widehat q-q_{\mathrm{ext}}\|_{L_2(\mu)}
+\|q_{\mathrm{ext}}-q^\star\|_{L_2(\mu)}\\
&=\sqrt{\epsilon_{\mathrm{fit}}}+\sqrt{\epsilon_{\mathrm{ext}}}.
\end{aligned}
\end{equation}
Substitution proves \eqref{eq:candidate_regret}.
\end{proof}

% Queue the supplementary figure before the next page begins.
\begin{figure*}[!t]
\centering
\captionsetup[subfigure]{font=small,position=top,justification=centering,skip=4pt}
\begin{subfigure}[t]{0.30\textwidth}
\centering
\caption{Safety-aware ordering}
\includegraphics[width=\linewidth]{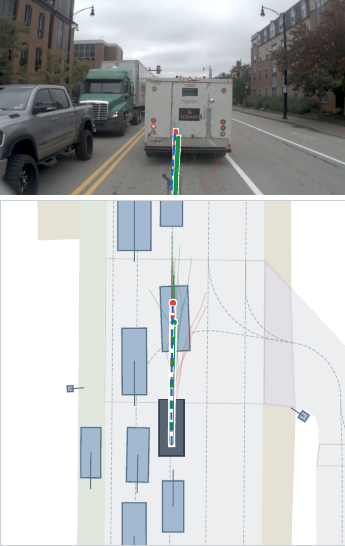}
\includegraphics[width=\linewidth]{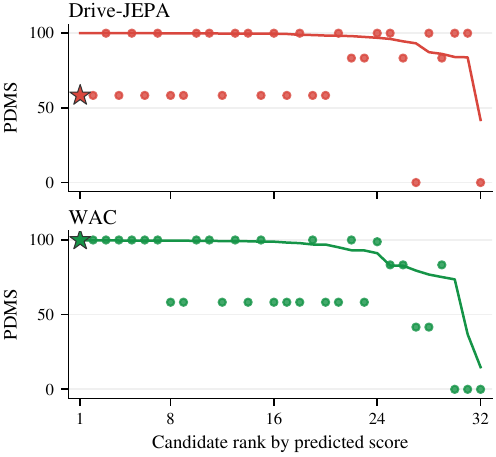}
\end{subfigure}\hfill
\begin{subfigure}[t]{0.30\textwidth}
\centering
\caption{Avoiding an overrated selection}
\includegraphics[width=\linewidth]{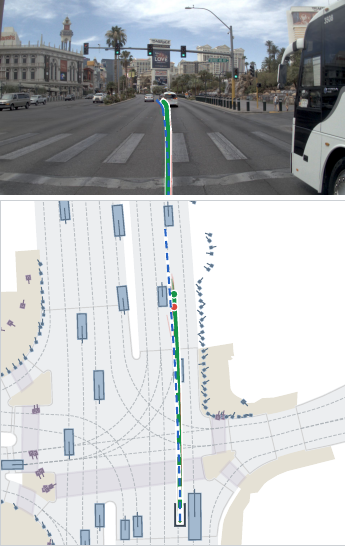}
\includegraphics[width=\linewidth]{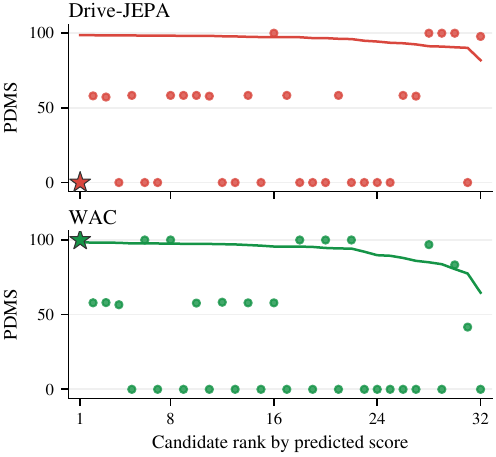}
\end{subfigure}\hfill
\begin{subfigure}[t]{0.30\textwidth}
\centering
\caption{Progress--safety tradeoff}
\includegraphics[width=\linewidth]{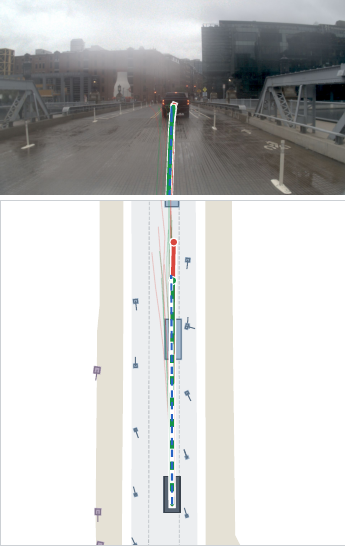}
\includegraphics[width=\linewidth]{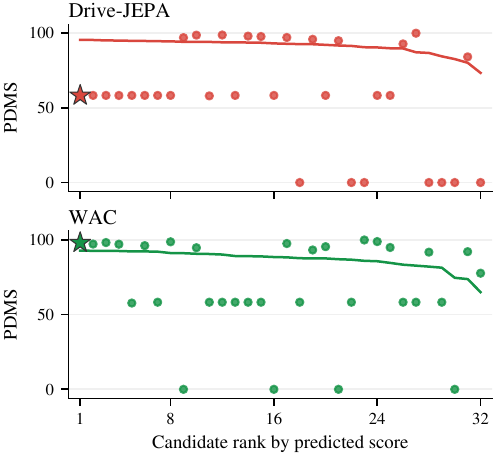}
\end{subfigure}
\par\vspace{3pt}
\includegraphics[width=0.98\textwidth]{figures/reselected/wacm_legend.pdf}
\caption{Candidate evaluation and planning decisions. Thin trajectories show candidates and thick trajectories their selections. Below each scene, candidates are sorted by descending predicted PDMS within each method. Curves show predictions, dots evaluator scores, and stars the evaluator score of the selected candidate. Scores use the 0--100 scale.}
\label{fig:wacm_additional}
\end{figure*}

\paragraph{Connecting critic learning to selection.}
For squared loss and square-integrable evaluator targets, conditional expectation gives
\begin{equation}
\mathbb E(\widehat q(z,a)-q)^2
=\epsilon_{\mathrm{fit}}+\mathbb E_\mu\operatorname{Var}(q\mid z,a).
\end{equation}
Thus $\epsilon_{\mathrm{fit}}$ equals the excess population loss over the optimal measurable predictor. The evaluator discrepancy $\epsilon_{\mathrm{ext}}$ separately accounts for alignment between the supervision and the intended task score.

\paragraph{Coverage and the proposed action set.}
The coverage condition applies to the joint test latent-state--action distribution. It transfers critic accuracy on supervised pairs to the candidates encountered during selection. The theorem bounds regret relative to the best proposed action. Comparing with the optimum over all feasible actions additionally includes the proposal gap, separating the role of the policy in generating candidates from the role of the critic in selecting among them.

\raggedbottom
\section{Additional Qualitative Results}
\label{sec:additional_cases}

Figure~\ref{fig:wacm_additional} pairs planning trajectories with candidate rankings in three urban driving scenes.

\paragraph{Safety-aware ordering.}
In Fig.~\ref{fig:wacm_additional}(a), WAC selects a shorter trajectory behind the truck that passes TTC while retaining the baseline's evaluator progress reward. The baseline's preferred candidates mix TTC failures with safer alternatives, whereas WAC ranks high-quality candidates first.

\paragraph{Avoiding an overrated selection.}
In Fig.~\ref{fig:wacm_additional}(b), the baseline's highest-scored trajectory fails the simulated collision check, while WAC's selection passes both collision and TTC checks. The selected paths have similar lateral endpoints but different outcomes.

\paragraph{Progress--safety tradeoff.}
In Fig.~\ref{fig:wacm_additional}(c), WAC achieves TTC compliance with a shorter forward trajectory and a small reduction in evaluator progress relative to the baseline. The baseline favors greater progress despite TTC failures, while WAC's leading candidates combine substantial progress with better temporal safety.

\section{Candidate Scoring Evaluation Details}
\label{sec:scoring_protocol}

\subsection{Evaluation Protocol}

We compare the complete planners on the NAVSIM-v1 test split, evaluating each on its own full candidate pool and excluding recorded trajectories. Predictions are the aggregate PDMS scores used for selection, without test-set calibration. The official simulator and scorer evaluate each candidate together with the PDM reference, preserving the official progress normalization independently of other proposals.

\subsection{Ranking and Overestimation Metrics}

\paragraph{Candidate ordering.}
Spearman correlation measures agreement between predicted and evaluator rankings, using average ranks for ties and excluding scenes where either ranking is constant. Pairwise ranking accuracy measures whether candidates with distinct evaluator scores are ordered correctly, with half credit for tied predictions. Each metric is computed within a scene and averaged equally across valid scenes. These metrics measure ordering rather than absolute calibration.

\paragraph{Low-score overestimation.}
For candidates with evaluator PDMS below 50 on the 0--100 scale, we measure the positive prediction error $\max(p-y,0)$, where $p$ and $y$ are the predicted and evaluator scores. We average within each scene and then equally across scenes where both methods produce such candidates.

\subsection{Uncertainty Estimation}

Uncertainty is assessed by paired bootstrap resampling of driving logs, keeping scenes from the same log together to account for their dependence.

\section{Implementation and Training Details}
\label{sec:implementation_appendix}

Table~\ref{tab:wacm_implementation} summarizes the architecture and training hyperparameters.

Factual future targets are computed with the shared visual and scene encoders under stop-gradient. The BCE configuration weights the six critic outputs equally. For the focal-weighted variant, let $y_c\in[0,1]$ denote an evaluator target, $p_c$ its predicted probability, and $p_{t,c}=y_cp_c+(1-y_c)(1-p_c)$. For NC, DAC, TTC, and comfort, the component loss is
\begin{equation}
\mathcal{L}_{\mathrm{focal},c}
=(1-p_{t,c})^{\gamma}\operatorname{BCE}(p_c,y_c),\qquad \gamma=2.
\label{eq:component_focal}
\end{equation}
EP and aggregate PDMS retain ordinary BCE.

\begin{table}[H]
\centering
\caption{Key architecture and training hyperparameters.}
\label{tab:wacm_implementation}
\small
\setlength{\tabcolsep}{4pt}
\renewcommand{\arraystretch}{1.08}
\begin{tabularx}{\columnwidth}{@{}>{\raggedright\arraybackslash}p{0.51\columnwidth}X@{}}
\toprule
Parameter & Value \\
\midrule
Visual backbone & ViT-L, 24 layers \\
Joint world--critic layers & 2 \\
Trajectory decoder hidden layers & 2 \\
Model / FFN hidden width & 256 / 1024 \\
Attention heads & 8 \\
World--critic activation & SwiGLU \\
Optimizer & RADAR \\
Learning rates & $10^{-5}$ (encoder), $10^{-4}$ (others) \\
Per-device batch size & 8 \\
Training epochs & 20 \\
\bottomrule
\end{tabularx}
\end{table}

\section{Acknowledgements}
This work is supported by Didi Voyager Research Labs, and also supported by NSF China with 92582205, NSF China with 52602738.